\documentclass{article}
\usepackage{PRIMEarxiv}

\usepackage[utf8]{inputenc} 
\usepackage[T1]{fontenc}    
\usepackage{hyperref}       
\usepackage{url}            
\usepackage{booktabs}       
\usepackage{nicefrac}       
\usepackage{microtype}      
\usepackage{lipsum}
\usepackage{fancyhdr}       
\usepackage{graphicx}       
\usepackage{amsmath,amssymb,amsfonts,upref,endnotes}
\usepackage{amsthm}
\usepackage{soul}
\usepackage{ragged2e}
\usepackage[skins]{tcolorbox}
\usepackage{dashrule}
\usepackage{capt-of}

\usepackage[export]{adjustbox}
\definecolor{jobcolor}{cmyk}{1,.10,0,.10}
\definecolor{joblightcolor}{cmyk}{0.10,0.068,0,0.012}
\usepackage{amsmath}

\title{Artificial intelligence for science: The easy and hard problems}

\author{
  Ruairidh M. Battleday and Samuel J. Gershman \\
  Department of Psychology and Center for Brain Science \\
  Harvard University \\
  \texttt{\{battleday, gershman\}@g.harvard.edu}
}

\begin{document}
\maketitle

\begin{abstract}
 A suite of impressive scientific discoveries have been driven by recent advances in artificial intelligence (AI). These almost all result from training flexible algorithms to solve difficult optimization problems specified in advance by teams of domain scientists and engineers with access to large amounts of data. Although extremely useful, this kind of problem solving only corresponds to one part of science---the ``easy problem.'' The other part of scientific research is coming up with the problem itself---the ``hard problem.'' Solving the hard problem is beyond the capacities of current algorithms for scientific discovery because it requires continual conceptual revision based on poorly defined constraints. We can make progress on understanding how humans solve the hard problem by studying the cognitive science of scientists, and then use the results to design new computational agents that automatically infer and update their scientific paradigms.
\end{abstract}



\section*{The easy problem}
Most work applying AI to science has focused on what might be called the ``easy problem.'' This is a relative term, since the easy problem is actually quite hard. A scientist specifies a function that they want to optimize (e.g., a function that generates a protein's structure given its amino acid sequence). Included in the specification is the input for the function (e.g., the amino acid sequence), the output (e.g., the 3D structure), and a way to compare the function's output with the ground truth or a desired range of outputs (e.g., the average 3D distance of an amino acid residue from where it should be). The scientist then finds or collects a dataset, usually very large, with examples of the ground truth, and applies AI optimization tools to the problem. So far, this kind of application has been highly successful, with new discoveries of tertiary protein structures, antibiotics, and nuclear fusion reactor designs (see \cite{wang2023scientific} for a review).

What makes this problem ``easy'' is not the form of the solution (which may require a great deal of engineering work) but rather the form of the problem. It is clear from the beginning what needs to be optimized, and what kinds of tools can be brought to bear on this problem. The engineering breakthrough comes from building much better versions of these tools. In other words, the problem is relatively easy because it does not require any conceptual breakthroughs of the sort involved in the discovery of relativity theory, genetics, or the periodic table.

\section*{The hard problem}
The fundamental barrier to automating science is conceptual. Great scientists are not simply extraordinary optimizers of ordinary optimization problems. It is not like Einstein had a better function approximator in his brain than his peers did; or Mendeleev’s brain had a better version of backprop. More commonly, great scientists are ordinary optimizers of extraordinary optimization problems. It is the formulation of the problem, not its solution, that is the truly hard problem: The hard problem is the ``problem problem.''

One might be tempted to relegate the hard problem to the fringes of ``revolutionary science,'' which rarely erupt into mainstream scientific practice, whereas the easy problem occupies the focus of the ``normal science'' that scientists spend most of their time on \cite{kuhn1962structure}. However, normal science is not simply optimization. This is obvious to any first-year graduate student trying to figure out what to work on. Normal science isn’t a catalog of optimization problems waiting to be solved by a queue of grad students. Their fundamental barrier is the same one facing AI scientists: It is the conceptual problem of formulating an optimization problem. This encompasses both major conceptual breakthroughs, like relativity theory, and the more modest ones achieved by graduate students on a regular basis, which nonetheless remain out of reach for existing AI systems.

\section*{The problem with optimization}
Much of the classic work on AI science (mainly by Simon, Langley, and their collaborators \cite{simon1981scientific}; but also more recently \cite{schmidt09lipson, udrescu20tegmark, sindy16brunton, jumper2021highly}) focuses on the easy problem. For Simon and Langley, this approach was premised on the psychological thesis that scientific cognition was essentially the same as regular problem solving, only applied to a different (and sometimes more challenging) set of problems. Consequently, they developed algorithms that emulated human problem solving, and applied these to model seminal scientific discoveries, including the (re-)discovery of oxygen with STAHLp \cite{rose1986chemical}. More modern methods for automated physics have also inferred many existing and novel laws, including classical and quantum problems with AI Feynman and non-linear dynamical systems with SINDy \cite{schmidt09lipson, udrescu20tegmark, sindy16brunton}; and, discovery algorithms in biology have advanced our ability to solve many difficult problems, including AlphaFold2 for protein folding \cite{jumper2021highly}.

This success is analogous to the earliest use of computers, in which they were used to complete calculations too laborious for any human. Algorithms that solve the easy problems of science are useful, even essential, to progress. For example, there is an increasing discrepancy between the number of amino-acid sequences that are discovered in biology and the recovery of the 3D protein structures they correspond to using the experimental method.

While recognizing the importance of such algorithms, we should also recognize their limitations. Several decades ago, Chalmers, French, \& Hofstadter challenged the idea that this kind of optimization was a complete model of scientific discovery and investigation \cite{chalmers1992high}. Systems like STAHLp are only able to solve scientific problems and make discoveries, they argued, because the modelers have \textit{represented the inputs and outputs to the problem in hindsight}. Only relevant data have been included and those data are already organized such that the proposed heuristics will be able to easily extract the right solution. In other words, most AI scientists have been provided a representation of the scientific problem that already includes the basic primitives needed for the final theory, but skirt the central problem of representation itself: Where do the primitives come from, and how do we know if we have discovered the right ones? 

Simon insisted (contra Popper \cite{popper1959logic}) that there \textit{was} a logic of scientific discovery, but Simon's proposal was really a logic of scientific problem solving---how to sequentially search through hypotheses given a problem statement and primitive representations \cite{simon1973does}. This is not discovery in the sense of problem creation. The latter involves representation learning in service of the problem, but also something deeper: Identification of the goal or objective function itself. 

\section*{Solving the hard problem}
In contemplating how to build AI systems that solve the hard problem, it is instructive to look at how human scientists do it. At a high level, humans break the hard problem down into two sub-problems:
\begin{itemize}
    \item Domain specification. What are the relevant phenomena that need to be explained by a theory?
    \item Constraint specification. What kinds of constraints need to be imposed on a theory based on existing knowledge (both domain-specific and domain-general)?
\end{itemize}
Once the domain and constraints have been specified, we can define an optimization problem (theory search); hence, we have converted the hard problem into the easy problem. However, it is uncommon for real scientists to do a single pass from hard to easy, because they often realize that the problem they are solving is the wrong one. This may happen for several reasons. One is the realization that a theory is internally inconsistent or paradoxical. Another is the realization that the theory may (with suitable modification) be able to explain a broader range of phenomena, prompting a respecification of the domain. Conversely, phenomena which were previously included in a domain may need to be excluded if no adequate unifying theory is found for all the phenomena. Respecification can also happen when new empirical phenomena are reported. In a related vein, constraint respecification can happen when domains are merged, split, expanded, or shrunk. 

The key point is that problem creation and problem solving are cyclically coupled in scientific practice. For most current AI scientists, on the other hand, the modeling team has already conducted domain and constraint specification once, in advance---in the representation and selection of data, and the in the representational scheme for potential scientific theories (outputs) and the objective function that assesses them, respectively.

In the following sections, we motivate the distinction between the easy and hard problems with three case studies from the birth of modern chemistry, physics, and molecular biology. For each case study, we summarize the elements of the problem, the historical setting, and modern computational systems that have tried to recapture some aspects of these discoveries. We will argue that none of these systems offers a complete solution to the hard problem.



\section*{Case study 1: The discovery of oxygen}
In the 18th century, it had been observed that lead increased in weight when it was slowly heated (which today we call ``oxidation,’’ but at that time was called ``calcination’’). This was difficult to explain with contemporary chemical theories, because they posited that something \textit{left} a metal when it was heated (a type of inflammable earth called ``phlogiston''). In 1774, the English chemist Joseph Priestley collected and identified a particularly inflammable and respirable form of air following the thermal reduction of calx-of-mercury (mercury-oxide) \cite{priestly1775Oxygen}. The French chemist Antoine Lavoisier eventually called this air ``oxygen,’’ and posited that it went \textit{into} the metal during calcination, causing the weight change \cite{lavoisier1790kerr}. Lavoisier's course of investigations were so successful that he has been credited as having started the Chemical Revolution and introduced the principled application of the conservation of mass into the quantitative sciences.

\subsection*{STAHLp}
Rose and Langley proposed a computational model called STAHLp to account for the discovery of the role of oxygen in calcination reactions \cite{rose1986chemical} (see Box 1). The input to STAHLp is a set of interconnected beliefs about 1) which substances are present before and after a particular reaction or 2) the chemical composition of each substance. These inputs are encoded using two types of variable: The predicates (or programs) REACTS and COMPOSED OF, and chemical names which the functions operate over. STAHLp's desired output is a consistent ``theory''---a set of beliefs that entail the inputs and do not contradict each other.

STAHLp solves this problem by applying a set of ``production rules'' to its beliefs, generating further beliefs until no more rules can be applied. If the system generates an inconsistent belief, STAHLp throws an error. At that point, a second set of ``belief revision'' heuristics is applied to try to identify the source of the inconsistency and correct it. After the lowest-cost correction is made, STAHLp applies its production rules to generate the updated theory entailed by the new starting beliefs. This is what Simon meant by scientific problem-solving as search through a hypothesis space \cite{simon1973does}.

\begin{figure}[h]
    \centering \includegraphics[width=0.75\linewidth, trim={0 0 0 0}, clip]{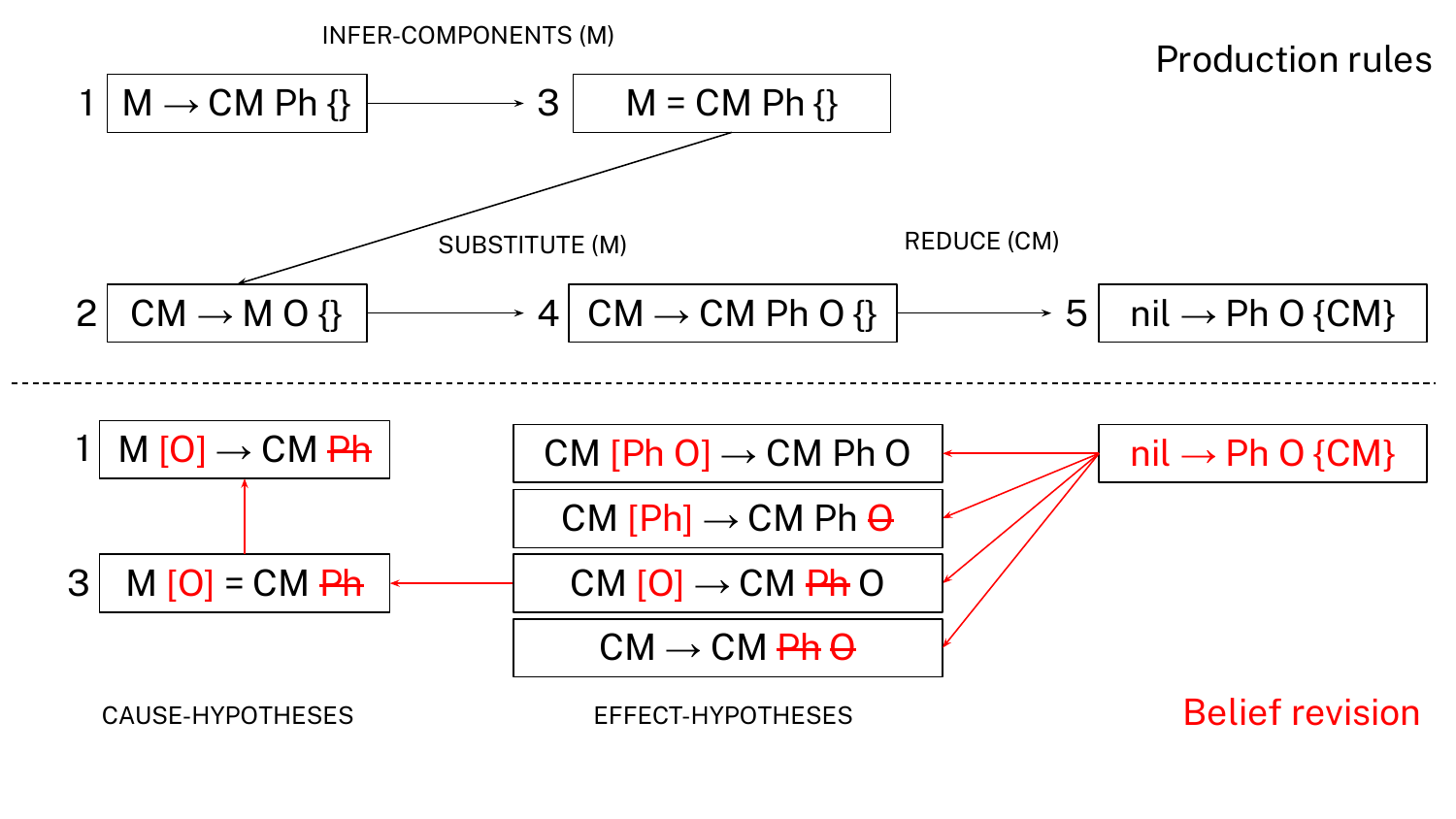}
    \caption{The discovery of oxygen by STAHLp. The input is two beliefs, where the first belief, inherited from Georg Ernst Stahl, states that mercury (M) is composed of calx-of-mercury (CM) and phlogiston (Ph). The second, reflecting an empirical observation by Joseph Priestley, states that calx-of-mercury is composed of mercury and a colorless gas (O). Through forward chaining the system reaches a circularity present in the initial beliefs---that mercury-calx can be decomposed into itself, phlogiston, and oxygen. The belief revision rules are then triggered, leading to a set of effect hypotheses, then cause hypotheses, giving the revisions needed to overcome the circularity. Reproduced from \cite{rose1986chemical}.}
    \label{fig:stahlp}
\end{figure}

Rose and Langley showed that for a particular pair of beliefs STAHLp ``discovers'' oxygen (see Figure \ref{fig:stahlp}). When STAHLp's production rules are applied to these observations, they produce an inconsistent belief, which in turn triggers its update-belief rules, generating a set of ``effect hypotheses'' that ``balance'' the inconsistent belief  (belief \textbf{4} in Figure \ref{fig:stahlp}). By back-tracing where the left and right sides come from, STAHLp can generate ``cause-hypotheses'' about how the initials beliefs should be updated to correct the inconsistency. The cause-hypothesis that affect the least downstream statements is chosen---in this case the addition of ``oxygen'' to the left hand side of belief \textbf{1}.

\subsection*{Lavoisier}
The modeling choices behind STAHLp neatly illustrate the easy vs hard distinction. It is hard to argue with the assumption that reactions and compositions of substances were the central concepts in chemistry---indeed, this was how Stahl himself defined its scope \cite{stahl1723}. However, leading up to the chemical revolution, chemists had a different way of thinking about the internal structure of substances, in which observable substances arose from mixtures of the latent primitives of earth, water, and sometimes fire. Notably, most continental European chemists would not have considered air---what we now call gas---to have chemical properties and enter into chemical combinations \cite{guerlac1961lavoisier}. By respecifying the ``definitions of chemistry'' to include air, Lavoisier was able to include measurements about gross changes in air volume in his theories, which in turn explained the gross weight changes during combustion, calcination, and reduction by the chemical fixation or release of air. This kind of conceptual innovation, through the revision of hierarchical ontological constraints, has been implicated at the core of human conceptual development \cite{keil1979semantic}. 


\begin{tcolorbox}
[colback=joblightcolor,colframe=jobcolor,center title,title=Box 1: STAHLp, label={box:stahlp}] 

\centering{\bf Beliefs}
\vspace{-1mm}
\flushleft The inputs and outputs of STAHLp are two types of belief:
\newline {\bf Componential model:} ``Substance X is COMPOSED OF \{Y, Z\})'';\newline X = Y, Z
\newline
{\bf Reaction:} ``\{W, X\} REACT to produce \{U, Z\})''; \newline W, X $\rightarrow$ U, Z
\tcblower
\centering{\bf Production rules}
\vspace{-1mm}
\flushleft
STAHLp applies a set of production rules to generate further beliefs:\\
\vspace{1mm}
\begin{minipage}[t]{0.3\textwidth}
\centering
    \textbf{Substitute}:\newline
    W, X $\rightarrow$ U, Z\newline
    X = Y, Z \newline
    $\therefore$ W, Y, Z $\rightarrow$ U, Z
\end{minipage}
\hfill
\begin{minipage}[t]{0.3\textwidth}
\centering
\textbf{Reduce}:\newline
 W, Y, \st{Z} $\rightarrow$ U, \st{Z}
\end{minipage}
\hfill
\begin{minipage}[t]{0.3\textwidth}
\centering
\textbf{Infer-components}:\newline
U = W, Y
\end{minipage}
\tcbline
\centering{\bf Objective}
\vspace{-1mm}
\flushleft
The objective function STAHLp uses to assess the consistency of a theory:
\vspace{-1.5mm}
\begin{equation*}
    \begin{cases}
    \text{Fail} \text{ if } nil \in \text{production rules}(\text{beliefs})\\
    \text{Continue} \text{ otherwise.}
    \end{cases}
\end{equation*}
\vspace{-3.5mm}
\tcbline
\centering{\bf Belief revision rules}
\vspace{-1mm}
\flushleft 
If an inconsistent belief is generated, STAHLp applies a different set of belief revision rules to the beliefs upstream of the problematic statement. 
\end{tcolorbox}
\vspace{2.5mm}

Next, Lavoisier broadened his scope to include all operations that fix or release air \cite{holmes1997antoine}, with the aim of tracing the flow of air and water through different coupled reactions in order to infer the chemical composition of a more complex substance (like chalk). This richer set of data about weight and volume changes led to the development of quantitative models based on tables and rudimentary equations. Prior to Lavoisier, chemists categorized and weighed solids and liquids before and after reactions, but did not routinely measure the air surrounding these materials. So, their ``equations’’ seldom balanced, and the conservation of mass was used more as a post-hoc and abstract principle rather than a tool for quantitative purposes. In addition, in the 18th century it was widely held that substances could dissipate away to nothing---from diamond, to phlogiston itself \cite{guerlac1961lavoisier}.

Discrepancies in subsequent experiments led Lavoisier to the conclusion that there must be different \textit{subtypes} of air with different densities. This led to the development of new equipment to measure those densities, and ultimately the finding that the air of the atmosphere was in fact a composite of these subtypes, rather than an elemental root. He then showed that the reduction of calx-of-mercury with charcoal produced a different air (carbon monoxide and carbon dioxide) than the reduction of calx-of-mercury without charcoal, eventually calling the latter air ``oxygen.'' Lavoisier explained the differences between these two reactions by positing an underlying, potentially infinite range of chemical primitives that could take the familiar three states of matter depending on how much of the ``matter of fire'' was coupled with them. These primitives were simply those that could be isolated by the tools of the chemistry at the time. This was the beginning of the main Chemical Revolution---actually more of an inversion, in that the things previously considered elemental (earth, water, air, fire) were now considered complex, whereas previously complex things like carbon were now considered elemental. 

With this historical framing, the it is easier to see how STAHLp's specification of the problem is essentially a modern one. The objective function for STAHLp is based on the detection of nil statements, whereas Lavoisier had to develop the conceptual machinery---the right representation of the problem---to represent a reaction in terms of the total weight of materials at the start and end, and in doing so established the loss function to be optimized---the inference of a consistent and useful set of equations. In other words, he constructed the right representation of the problem. This placed emphasis on the use of a density constant to relate changes in air volume to changes in weight. Furthermore, a new chemical name could not be added arbitrarily, and had to be placed within the existing ontological structure that initially did not include air. ``Oxygen'' is simply not a valid entry \cite{guerlac1961lavoisier}. Finally there is also the question of data selection. The creators of STAHLp include only two facts from the many heterogenous and often inconsistent observations and beliefs in 18th century chemistry. If they took into account others---for example, that when nitrous acid was poured on mercury colored vapors and fumes were given off---the model's conclusions may well have changed. And, there was actually a great deal of inconclusive or even negative evidence that metals apart from lead increased weight on calcination, detracting from the general statement that an air entered into metals.




\section*{Case Study 2: The electromagnetic field}
In the mid-19th century, Michael Faraday published a set of discoveries and observations related to electromagnetic induction. Faraday recorded the intensity of current induced in a copper wire by moving it around magnets of various shapes, strengths, and number, as well as other electrical circuits embedded in magnetic media, arguing that the most useful representation for these data was in terms of \textit{lines of magnetic force} \cite{faraday1852lines}. He had speculated on what might be the cause of these patterns, but had been largely unsuccessful \cite{faraday1852physical}. The Scottish physicist James Clerk Maxwell derived a brilliant and creative theoretical solution to this problem that provides the foundation of modern physics---the mathematical representation of the electromagnetic field.

No computational model has been proposed to emulate Maxwell’s discovery. However, several influential models target the general setting of deriving physical laws from datasets of this sort \cite{schmidt09lipson, udrescu20tegmark, sindy16brunton}. Here we will focus on AI Feynman \cite{udrescu20tegmark}, an algorithm that uses \textit{symbolic regression} to recover natural laws from physical data (see Box 2).

\subsection*{AI Feynman}
AI Feynman takes a data table, comprising data samples (rows) of a dependent variable and several independent variables (columns), and outputs a symbolic formula representing a theory of the system as well a set of predictions in the input space. Variables take continuous values, correspond to measurements of the physical system, and are augmented with type information representing their fundamental physical units (meter, second, kilogram, kelvin, and volt). In order to infer the theory of how the independent variables determine the dependent variable, AI Feynman cycles through a set of pre-specified computational strategies premised on commonalities in the functional forms of solutions to known physical problems (Figure \ref{fig:Feynman}). The algorithm stops when the squared-error loss between its predictions and the input is low enough, and then checks whether its current solution is equivalent to the ground-truth expression. 

For example, the data in ``mystery table 5'' comprises samples from one dependent variable, $F$, and nine independent variables corresponding to the masses and 3D positions of two objects, and Newton's constant $G$. The algorithm runs through its pre-determined steps: Algebraic manipulations yield a reduced set of dimensionless variables; the application of a neural network component identifies translational symmetry; a good factorization is found; then polynomials are fit to two subsets of transformed variables. The end result of this process is an equation that accounts for the data below some error threshold, $\epsilon$ (see Figure \ref{fig:FeynmanTr}).

\subsection*{Newton and Maxwell}
Although the input variables chosen \textit{for each problem} in AI Feynman might seem logical, they in fact correspond to quite an advanced stage of problem solving---when scientists have already constructed an idealized model for the system at hand.\footnote{This is actually the process that Richard Feynman goes through in his lectures when giving the historical background of the problem statement.} For example, Newton had to \textit{posit} the idea of a gravitational constant, expressed implicitly in terms of proportionality; and he had to posit that these quantities were the \textit{only} influential factors when explaining gravity---that action-at-a-distance was the correct framework to use, rather than the transmission of forces through an underlying medium. 

Nancy Nersessian has given a thorough cognitive-historical analysis of Maxwell and the development of the electromagnetic field concept \cite{nersessian2010creating}. A striking feature of Maxwell's problem-solving is how explicit he was about the scope of his theories and the utility of intermediate models. Selectively restricting the domain allowed him to identify which parameters or features of the intermediate model were essential, and an analysis of those features afforded selective expansion of the domain---a process Nersessian has called ``generic abstraction.''

Maxwell began by restricting his scope to Faraday's data on electromagnetic induction and lines of force, in order to make progress given the ill-defined and heterogenous state of electrical science at the time. First, he constructed a \textit{descriptive} mathematical model of Faraday's observations and theoretical postulations, based on the continuum mechanics of stresses in an underlying medium \cite{maxwell1855lines}. He then used this new representation of the data to design an explanatory dynamical model, iterating between adding to the range of phenomena he was considering, and adapting his intermediate model to accommodate them and generate new predictions \cite{maxwell1861lines, maxwell1864lines}. He began with magnetic phenomena, and showed that the constraints provided by his descriptive analysis could be fit by a vortex model. From this model he could calculate the magnetic force at any point in the medium by carrying over the system of equations describing the mechanical force that would be exerted in the vortex model and replacing mechanical variables with magnetic ones \cite{nersessian2002maxwell}. When he generalized this model to a \textit{medium} composed of these vortices, however, he found the model unsatisfactory because of the friction that would occur between adjacent vortices. This brought to mind the idle wheels interposed between rotating machine gears, from which he introduced the idea of idle-wheel particles to communicate between vortices. Idle-wheel \textit{particles} provided a good way to model electrical current, so his next step was to include electromagnetic phenomena. But this required the relaxation of the model to allow the particles to translate in conductive medium, and to rotate without generating any friction. Using the new model, he could bring in a set of equations to represent electrical current as the flux density of these particles, driven by the circumferential velocity of the vortices \cite{nersessian2002maxwell}. Maxwell continued this process of domain relaxation and model building to include electrostatic phenomena and the polarization of light. 

\begin{tcolorbox}
[colback=joblightcolor,colframe=jobcolor,center title,title=Box 2: AI Feynman, label={box:feynman}] 
\begin{minipage}[t]{0.5\textwidth}
\textbf{Stepwise objective}
    \centering
    \begin{equation*}
        \begin{cases}
            1 \text{ if } ||\frac{(\hat{f}(x) - x)^2}{n}||^{\frac{1}{2}} < \epsilon_{\, \text{step}}\\
            0 \text{ otherwise.}
            \end{cases}
\end{equation*}
\end{minipage}
\begin{minipage}[t]{0.5\textwidth}
\textbf{Final objective}
    \centering
    \begin{equation*}
        \begin{cases}
            \text{Pass} \text{ if SIMPLIFY}(\hat{f}-f)==0\\
            \text{Fail} \text{ otherwise.}
            \end{cases}
\end{equation*}
\hspace{2mm}
\end{minipage}
\\
AI Feynman assesses whether the Euclidean distance between predictions and data is small enough after each step, and whether its symbolic expression is the same as the ground truth law.
\tcblower
\begin{minipage}[t]{0.486\textwidth}
    \includegraphics[trim={1cm 0.115cm 1cm 0.2cm}, clip, width=0.9\linewidth,valign=c]{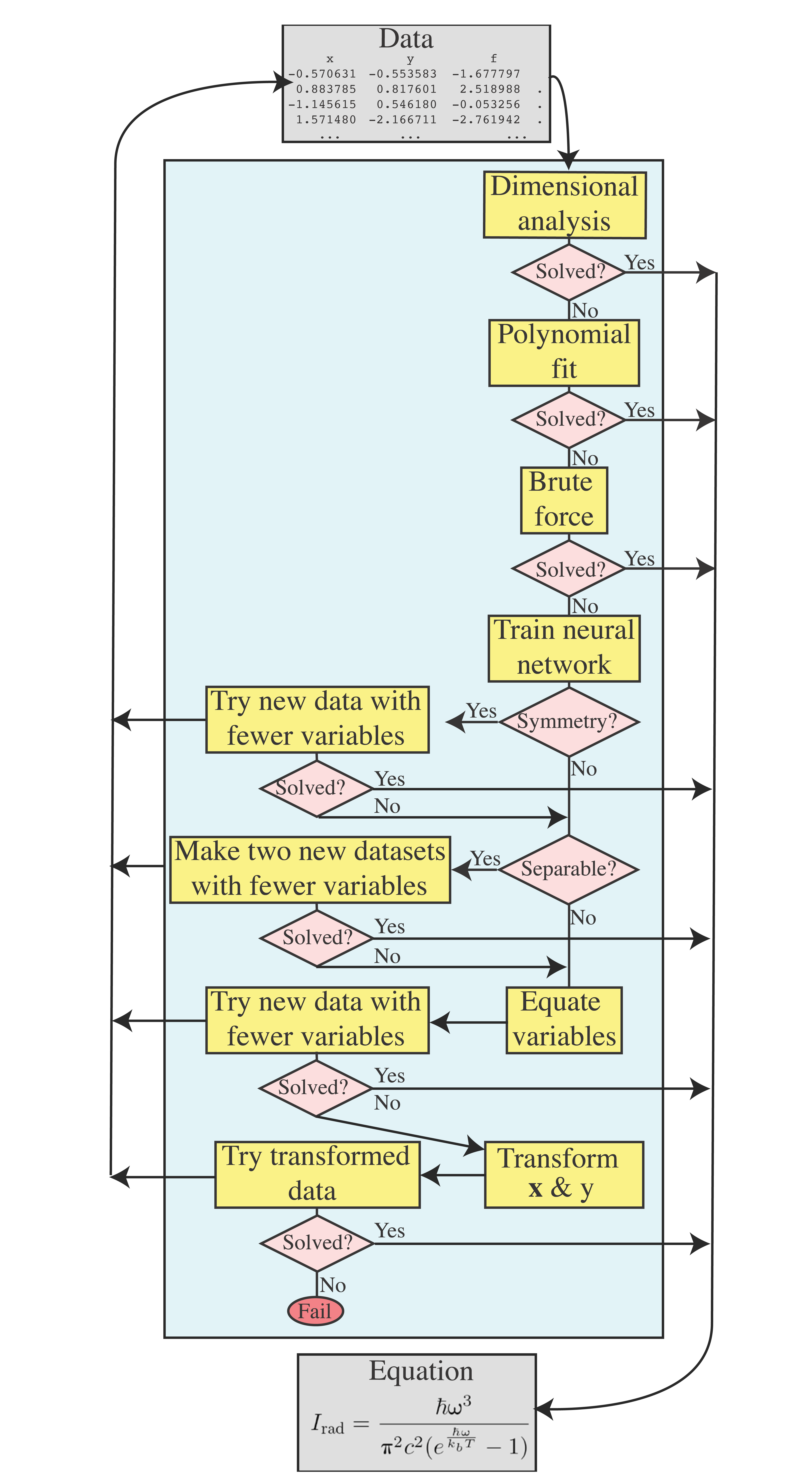}
\end{minipage}
\hfill
\begin{minipage}[t]{0.486\textwidth}
\centering
\includegraphics[width=0.85\linewidth,valign=c]{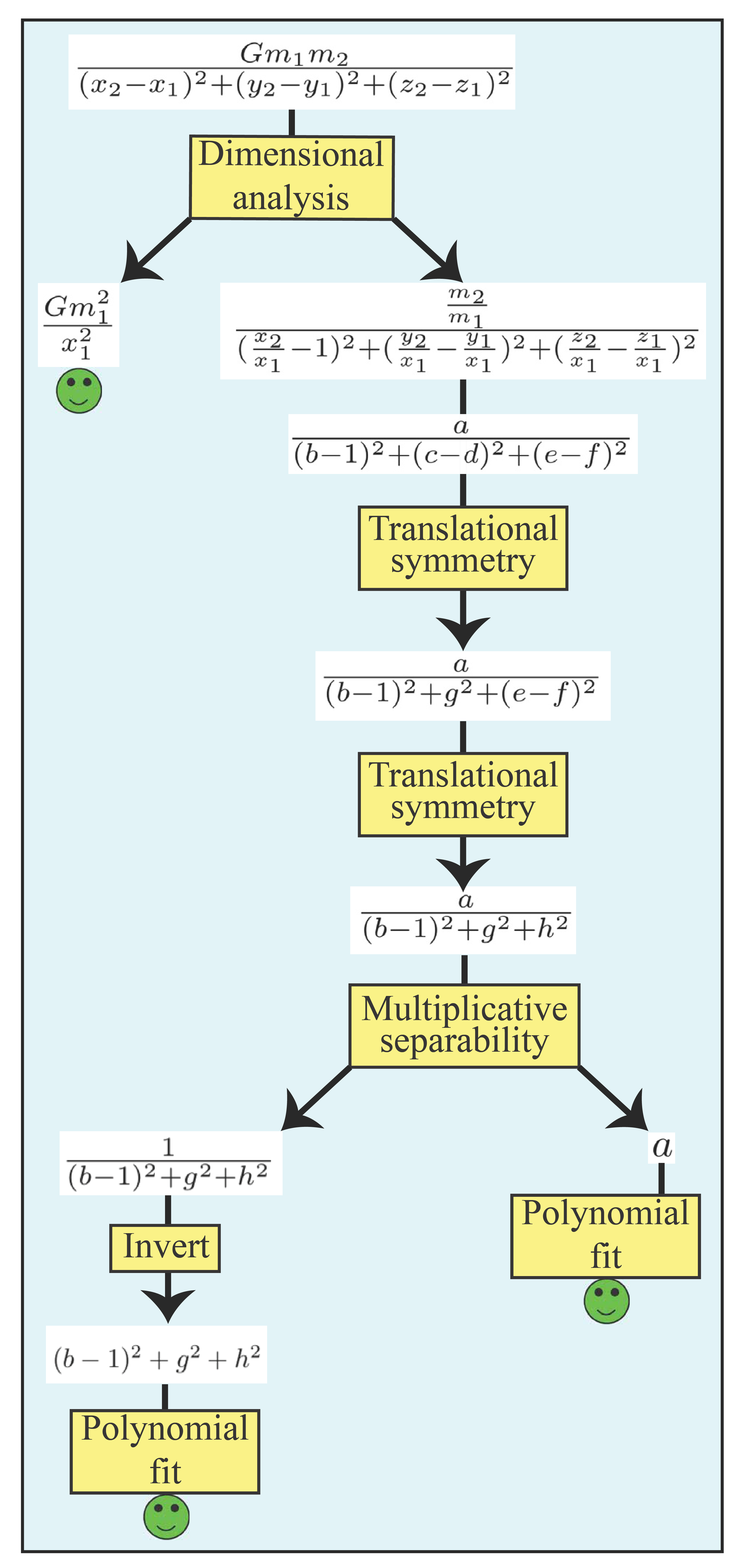}
\end{minipage}
\begin{minipage}[t]{0.486\textwidth}
\captionof{figure}{The steps that AI Feynman goes through when applied to solve a scientific problem, given in the form of a mystery table. Reproduced from \cite{udrescu20tegmark}.}\label{fig:Feynman}
\end{minipage}
\hfill
\begin{minipage}[t]{0.486\textwidth}
\centering
\captionof{figure}{AI Feynman recovers the correct expression at the top of the diagram by applying its problem-solving steps. Reproduced from \cite{udrescu20tegmark}.}\label{fig:FeynmanTr}
\end{minipage}
\vspace{-1mm}
\end{tcolorbox}

Like Lavoisier, Maxwell was guided in this process of abstraction by \textit{ontological knowledge} about the structure of different physical and mathematical systems, which also helped him sequentially assemble and modify the mathematical expressions underlying the model. Perhaps these idealized models played a role in Lavoisier's early investigations, albeit in a simpler form involving crude movements of air and changes of weight. What is clear is that this process is not captured by systems like AI Feynman, which are given the problem variables from the mature idealized model, and lack the flexibility to alter their own conceptual systems---analogous to providing the symbol $i$ before deriving a general solution to the problem of polynomial root finding. 

The kinds of laws AI Feynman can derive are also limited by its processing steps. This is motivated by an analysis of common characteristics of physical laws---they contain variables with units, low-degree polynomial structure, compositionality, smoothness, symmetry, and separability. But again, these constraints arose out of analysis of the existing laws of physics, and provide constraints that restrict the subsequent class of models in an inflexible manner---Maxwell \textit{invented} dimensional analysis to help solve difficult physics problems.

And, once again, there there is also the question of which datapoints are chosen. For mystery table 5, the data are not taken from systems far from the scientist or near large masses, where the behavior of light (its speed or deflection, respectively) needs to be taken into account. Recognizing and adjusting for these factors were essential parts of proving the theory and then taking it forward.

\section*{Case Study 3: Protein folding}
Several major conceptual breakthroughs led to the ``protein-folding problem.'' Frederick Sanger discovered that proteins are \textit{linear} chains of amino acids based on the isolation and recursive extraction of hydrolyzed fragments of insulin using various media and electrical currents \cite{sanger1951amino}. Evidence that the function of proteins depended on their 3D structure, rather than the identity of individual amino acids, came from X-ray crystallography \cite{kendrew1958three}, the structural effects of natural and artificial variation of amino acids \cite{perutz1965structure}, and catalytic-rate analyses with different cellular conditions, substrates, and inhibitors \cite{thoma1960competitive}. But it was Christian Anfinsen Jr. that put forward what became known as ``Anfinsen's dogma''---that under physiological conditions the primary sequence itself, and no other factors, determined the 3D structure \cite{anfinsen1973principles}.

\subsection*{AlphaFold2}
One of the most successful recent discovery algorithms is AlphaFold2 \cite{jumper2021highly}, which is a large neural network model that predicts the 3D structure of a protein given its 1D amino-acid sequence (see Box 3). When it was released, AlphaFold2 brought the average molecular deviation for a protein down from 0.3 to 0.1 nanometers, which was finally precise enough for biologists to make use of. AlphaFold2 uses complex heuristics to solve this optimization problem, based on a great deal of biological and engineering knowledge. At a high level, its \textit{Evoformer} module learns increasingly rich and abstract representations of the 1D primary structure {and} and a 2D matrix of inter-atomic distances between residues, and its \textit{Structure} module uses these representations to build a 3D model of the protein. The network is trained end-to-end, meaning all operations are differentiable and the loss signal from the final 3D positions is back-propagated to inform the update of neural networks weights in all operations after the input.

AlphaFold2's input is a multiple sequence alignment (MSA), which augments the protein of interest's 1D amino-acid sequence with additional rows containing similar amino-acid sequences from existing databases. If any of the MSA's sequences have already had structures derived, 2D distograms of the pairwise distance between residues and a sequence of torsion angles between adjacent amino acid residues are added to the inputs. AlphaFold2 outputs a set of atomic co-ordinates, a confidence score in each residue's position, torsion angles between adjacent amino-acid backbones, the 2D distogram between residues, and a prediction of any masked parts of the MSA. The objective function during training contains a loss term for each of these representations, with the most important components penalizing the 3D deviations of heavy atoms in the amino-acid chain. The loss function during ``fine-tuning'' contains all of these terms, plus two extra terms that penalize the final structure for violating physical constraints.

\subsection*{Anfinsen}
AlphaFold2's success comes in large part from the engineering choice of problem statement. In particular, it does \textit{not} solve the original ``problem'' of protein folding. 
Anfinsen and his colleagues began by studying bovine ribonuclease, a protein that could be isolated in abundance using the techniques of the time. Ribonuclease has a functional structure that depends on the (single) correct formation of four disulphide bonds (out of a possible 105 combinations between eight sulfhydryl groups). When these were broken, then allowed to reform in denaturing conditions (containing a high concentration of urea), an inactive mixture of all the possible configurations was generated. In physiological conditions, however, the mixture was converted to the native ribonuclease. Anfinsen and colleagues moved on to study Staphylococcal nuclease, which does not depend on disulphide bonds for its 3D functional structure, analyzing the folding and functionality of various sub-sections of the molecule as well as the temporal- and pH- dynamics of the rather sharp phase transition between inactive and active molecules and the initial few folding events in renaturation. The cumulation of these events was the ``thermodynamic hypothesis''---that the correctly folded protein occupied the minimum free-energy state in its natural cellular environment \cite{anfinsen1973principles}. 

\begin{tcolorbox}
[colback=joblightcolor,colframe=jobcolor,center title,title=Box 4: AlphaFold2, label={box:alphafold}] 

\centering
\textbf{Architecture}\\
\vspace{1.5mm}
\includegraphics[width=\linewidth]{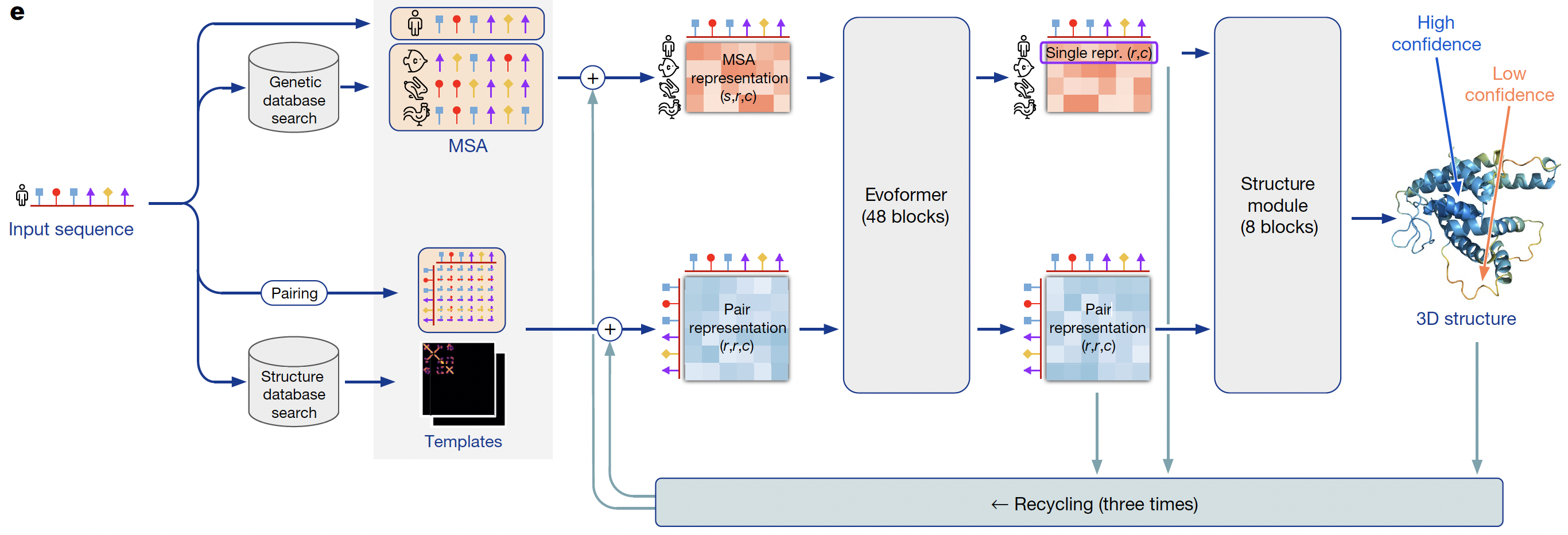}
\captionof{figure}{AlphaFold2 input is a 1D amino-acid sequence and any related sequences from genomic databases (the MSA), as well as distance information from previously derived structures. The Evoformer module learns flexible and increasingly abstract re-representations of these inputs over many layers. The Structure module conditions on this information to output rigid 3D frames for each amino acid residue, before adjusting the frames to form a physically plausible structure during fine-tuning. Reproduced from \cite{jumper2021highly}.}\label{fig:alphafold}
\tcblower
\centering
\textbf{Objective}\\
\vspace{1mm}
\includegraphics[width=\linewidth]{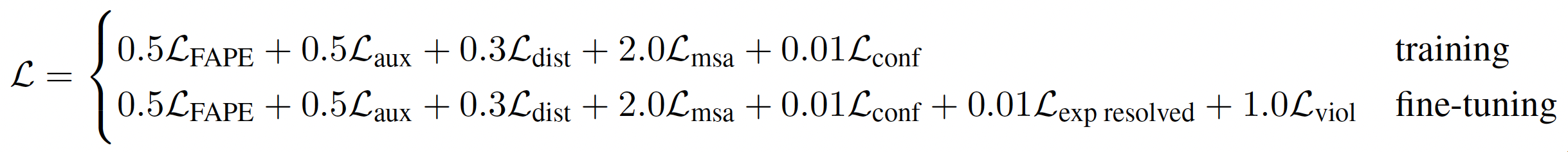}
\flushleft
AlphaFold2's training loss  contains terms for each representation maintained by the algorithm (the 3D positions of each amino acid residue, the torsion angles between amino acid residues, the pairwise distances from the 2D distogram, the masked 1D input sequence, and the model's confidence score). The fine-tuning loss has two extra terms that penalize the violation of physical constraints. Reproduced from \cite{jumper2021highly}.
\end{tcolorbox}

However, the task left facing scientists working on the protein-folding problem was daunting: Accounting for the full physical process by which a protein of $N$ amino acids assumed one of its $\sim 8^N$ possible conformations. Part of the genius of Alphafold2 was that its architects recognized that this problem did not need to be solved in order to make substantial impact in biology, and they relaxed its constraints to define a new, related problem: The prediction of a final folded state given the 1D sequence, leaving behind the requirement to model the time-evolving movement of the polypeptide chain from 1D denatured to 3D functional state. This respecification allowed them to bring in an abundance of sequence data, which can be used for the new, but not the old, problem. Evolutionary correlations have been used for some time to make arguments about folded structures and function \cite{fitch1970usefulness}, but do not obviously inform folding dynamics. It also allowed them to bring in the very flexible learning algorithms from modern deep learning. In particular, attention-based mechanisms model longer-range dependencies over 1D sequences, making use of the 2D distogram representations, and training could be split into a free optimization period, where physical constraints can be violated during iterative representation learning and stochastic gradient descent over the residue gas representation, followed by a bespoke fine-tuning stage where physical constraints were met.

The positive consequence of this choice is that some requirements of a solution to the original problem are met---we can predict the structure of hydrophilic proteins with lots of analogous evolutionary sequences well. The negative consequence is that we don't have a model of the folding dynamics which can make good predictions of the structures of orphan molecules like antibodies, lipophilic molecules with no experimentally derived homologous structures, or the effect of a new mutation or ion on the final folded structure.



Although the scale and complexity of natural biological systems warrants different types of theory and strategies of investigation, including decomposition and localization \cite{bechtel2010discovering}, there are also important commonalities---including the use of ontologies \cite{darden1991theory} and imagistic intermediate models \cite{nersessian2010creating, fitch1970usefulness}. Conversely, Lavoisier and Maxwell often re-specified problem constraints to bring in abundant data and a powerful method nearby.

 


\section*{Understanding the hard problem}
The previous sections depict a recurring pattern: Much progress in applications of AI to science has been made, but only with the aid of humans specifying the problem formulation. Thus, these systems are essentially solving the easy problem, not the hard problem. What makes the hard problem so hard?

An important and elusive feature of problem specification is that it is not a data modeling problem. The selection of what to model and and what constraints to condition on are antecedent to any data modeling problem. It is also not reducible to a representation learning problem, in the sense of figuring out how raw sensory input maps to abstract representations. Of course, that problem also needs to be solved, but first the scientist needs to know what problems the representations are being used to solve. Are these conceptual breakthroughs just patterns that can be discovered with a sufficiently powerful pattern recognition system? In a sense yes, but before that can happen, something has to tell the pattern recognition system what kind of patterns are interesting, important, and useful. What problem is the pattern-recognition system designed to solve, and where does this come from?

Sociological, aesthetic, and utility considerations enter at the problem specification stage. Building an AI scientist is as much about shaping its tastes, style, and preferences as it is about endowing it with powerful problem-solving abilities. Again, a look at how we train human scientists is instructive: A good graduate advisor educates students about what problems matter, what phenomena are interesting, which explanations count, and so on. These considerations can’t be brushed aside as subjective factors irrelevant to the purely technical problems facing AI systems; they are in fact constitutive of those technical problems. Without them, the technical problems would not exist.

A research program for attacking the hard problem should begin with the cognitive science of science \cite{thagard2012cognitive}, focusing on the understudied subjective, creative aspects discussed above and how they interact with the objective aspects of problem solving. In the study of science, there is a natural continuum through rare but large breakaway conceptual developments, best studied through cognitive-historical analyses \cite{nersessian2010creating, darden1991theory, battleday2025Lavoisier}; daily scientific problem solving, studied through observation of practising labs  \cite{nersessianvitro}; and large datasets of decisions in simplified problem-solving settings in online naturalistic games \cite{allen2024using}. We can tests any hypotheses about science that we derive from these analyses in large-scale online behavioural experiments, where, for example, related studies already provide evidence that humans construct simplified mental representations to plan \cite{ho2022people}, and that iterative model-based revision of problem statements as a critical part of deriving a successful scientific solution \cite{clement1994use}.

\section*{Towards scalable AI scientists that solve the hard problem}
Once we understand what human scientists are doing with enough precision that we can formalize their activities, we can try to leverage these insights to build scalable AI scientists. At least initially, it is unlikely that these will be stand-alone systems, but rather more like research assistants or first-year grad students: Curious agents with some technical competence but in need of expert guidance. This guidance can come in the form of natural language instruction, reading curricula, and demonstrations. The growth of models beyond this requires the examination and emulation of the communal aspects of science and related cultural institutions. Lab meetings, conferences, and presentations and discussions are ultimately the place where judgements on the quality of a scientific problem are made. 

The use of natural language processing for scientific discovery is at the heart of the recently proposed ``AI Scientist'' \cite{lu2024aiscientistfullyautomated}, arguably the first artificial scientist that addresses the hard problem. The algorithm itself is a carefully designed system of large language models (LLMs), prompting schemes, a coding assistant, and templates for papers and conference guidelines that autonomously updates machine-learning (ML) code in order to generate scientific papers. In its inner loop the AI Scientist is given access to the training, testing, and visualization code for a simple ML model and dataset, along with several suggestions of innovative changes to the code and the overall objective of reducing the model's loss on held-out data. Its outer loop requires that it generate a range of ideas in natural language format, check their novelty using the internet, apply several ideas, write a ML paper for each of the ideas that ran successfully, review the paper, then update the paper. The proposed system comprises a carefully designed interface of language models, prompting schemes, a coding assistant, and templates for papers and conference guidelines. From the examples presented, the innovative ideas that the AI Scientist generates are mostly decisions to split variables or processing pathways, add new model components or training metrics based on previously successful strategies in the literature, and combine any of the above that improve the final loss. A particularly impressive part of the work is the ability to implement these high-level conceptual changes in the code example, including producing useful visualizations. 

Whether this system and its successors can produce radically innovative discoveries remains to be seen. Do such systems replicate human strategies such as ontologically guided constraint respecification, producing and modifying intermediate models, and re-specifying the problem based on knowledge of adjacent rich sources of data and available models? Natural language is certainly capable of capturing some aspects of the ontological structure of knowledge, and multimodal models should be able to create and maintain imagistic intermediate models of the scientific phenomenon. We would like AI scientists that can likewise recognize when progress has been slow on a particular problem, but adjacent sources of data and powerful models promise fulfilment on an intersecting set of desiderata. We would also like them to recognize when and how to gather more useful data when there is a mismatch with the use case---as recent improvements on using AlphaFold2 to predict human structures have done.


On the other hand, many scientific developments, including those we have characterized above, come from a reflective consideration of either how to alter model constraints to capture anomalous data \cite{laudan77, darden1991theory}, or where an alteration of model constraints affects the domain, borne out over a course of successive investigations \cite{holmes1997antoine, nersessian2010creating}. Whether current large language models can capture this type of reflective continual learning and selective conceptual respecification will require further investigation \cite{mitchell2020crashing, hase2024does}. For now, humans remain the only intelligent system capable of solving the hard problem. We still have much to learn about building AI scientists by studying ourselves.


\section*{Acknowledgments}
{We are grateful to Nancy Nersessian, Melanie Mitchell, Giovanni Pezzulo, Frank Keil, Skyler Wang, George Davis-Smith, John Clement, and Tom Griffiths for helpful discussions. This work was supported by the Kempner Institute for the Study of Natural and Artificial Intelligence, and by the Schmidt Sciences Polymath Program.}

\bibliographystyle{unsrt}

\end{document}